 \documentclass[review,5p,times,twocolumn,authoryear]{elsarticle}

\usepackage{amssymb}
\usepackage{lipsum}
\usepackage[switch]{lineno}
\usepackage{graphicx}
\usepackage{xcolor}

\usepackage{todonotes}
\usepackage{xcolor}
\usepackage{listings}
\usepackage{graphicx}      
\usepackage{subcaption}
\usepackage{url}
\usepackage{hyperref}
\usepackage{makecell}

\usepackage{booktabs}
\lstdefinestyle{sparqlstyle}{
  basicstyle=\small\ttfamily,
  keywordstyle=\color{blue}\bfseries,
  identifierstyle=\color{black},
  stringstyle=\color{red},
  commentstyle=\color{green!50!black}\itshape,
  showstringspaces=false,
  language=SPARQL, 
  breaklines=true,
  breakatwhitespace=false,
  basicstyle=\footnotesize\ttfamily,
    postbreak=\mbox{\textcolor{gray}{$\hookrightarrow$}\space},
  breakindent=0pt,
  tabsize=2,
  numbers=left,
  numberstyle=\tiny\color{gray},
  frame=single,
  rulesepcolor=\color{gray!20},
  backgroundcolor=\color{gray!5},
  captionpos=b,
  morekeywords={SELECT,WHERE,FILTER,OPTIONAL,BIND,UNION,GRAPH,SERVICE,GROUP,BY,ORDER,ASC,DESC,LIMIT,OFFSET,HAVING,DISTINCT,REDUCED,AS,VALUES,NOT,EXISTS,MINUS,PREFIX,CONSTRUCT},
  morestring=[b]", 
  morecomment=[l]{\#}, 
}

\usepackage{tikz}
\usetikzlibrary{positioning, fit, backgrounds, arrows.meta, decorations.pathreplacing}

\definecolor{nutrient}{RGB}{30,100,160}
\definecolor{disease}{RGB}{46,125,50}
\definecolor{foundation}{RGB}{0,121,107}
\definecolor{integration}{RGB}{198,124,0}
\definecolor{datatier}{RGB}{84,110,122}

\journal{Computers and Electronics in Agriculture}

\begin{document}
\begin{frontmatter}



\title{From Field Data to Global Food Systems Intelligence: A Semantic Graph Framework for Sustainable Wheat Production}


\author[first]{Nirmal Gelal}
\author[second]{Aastha Gautam}
\author[first]{Soheil Abadifard}
\author[fourth]{Nico Giordano}
\author[first]{Moumita Sen Sarma}
\author[first]{Sanaz Saki Norouzi}
\author[second]{Claudio Dias da Silva Jr}
\author[second]{Jean Ribert Francois}
\author[first]{Kathleen M. Jagodnik}
\author[third]{Katherine Nelson}
\author[second]{Terry Griffin}
\author[second]{Xiaomao Lin}
\author[second]{Stacy Hutchinson}
\author[second]{Stephen M. Welch}
\author[second]{Kelsey Andersen Onofre}
\author[second]{Romulo Lollato}
\author[first]{Pascal Hitzler}
\author[first]{Hande Küçük McGinty\corref{cor1}}
\cortext[cor1]{Corresponding author, hande@ksu.edu}

\affiliation[first]{organization={Department of Computer Science, Kansas State University},
            addressline={}, 
            city={Manhattan},
            postcode={66506}, 
            state={Kansas},
            country={USA}}
\affiliation[second]{organization={Department of Agronomy, Kansas State University},
            addressline={}, 
            city={Manhattan},
            postcode={66506}, 
            state={Kansas},
            country={USA}}
\affiliation[third]{organization={School of Natural Resources, University of Missouri},
            addressline={}, er
            city={Columbia},
            postcode={65211}, 
            state={Missouri},
            country={USA}}

\affiliation[fourth]{organization={Department of Agronomy \& Horticulture, University of Nebraska-Lincoln},
            addressline={},
            city={Lincoln},
            postcode={68583},
            state={Nebraska},
            country={USA}}

\begin{abstract}
In response to the growing need for structured, interoperable agricultural data, this paper presents the Sustainable Wheat Production Datahub, a modular, graph-based framework that brings diverse wheat production datasets together into a single, queryable store. Using the Knowledge Acquisition and Representation Methodology (KNARM), with domain experts in the loop, we developed ontologies for nutrient management and disease management, two of the areas that most affect wheat yield and its sustainability, covering practices such as nitrogen fertilization and fungicide-based disease control. The two ontologies are federated, meaning they are maintained as separate but connected modules, joined by a bridging layer of  cross-domain links, and together they form the schema of a knowledge graph (KG). We construct this KG by populating the ontologies with data from diverse sources, including field trials, expert knowledge, and environmental descriptors. We validate the ontologies, showing the KG accurately answers practical agronomic questions expressed in  SPARQL. We demonstrate that the KG  derives facts entailed by the ontologies beyond those explicitly stored, and that a single query can draw across independently sourced datasets. The framework also has practical implications for wheat research and extension programs, since it makes data from different sources easier to find, combine, and reuse. Designed to expand toward the full wheat lifecycle, from farm to table, this work establishes the foundation for a scalable, semantically rich global food systems datahub.
\end{abstract}




\end{frontmatter}




\section{Introduction}
The growing need for structured, machine-readable representations of domain knowledge has become increasingly important. Ontologies, a formal and hierarchical data representation technique that defines the relationships between concepts within a specific domain, have become a functional tool in capturing the complexities inherent in data. They enable semantic understanding, interoperability, and automated reasoning across various fields. Similar to ontologies, Knowledge graphs (KGs) use ontological principles to store multi-relational data as a graph structure comprising Resource Description Framework (RDF) triples. KGs and ontologies, widely utilized in fields such as bioinformatics and healthcare \cite{GeneOntology,CARO,kohler2021human,lin2017drug}, are gaining popularity in agriculture to express complex domain knowledge and manage extensive domain vocabularies \cite{roussey2010ontologies,malik2015ontology}.

Wheat accounts for 20\% of calories and 11\% of protein in the human diet \cite{Shewry_2007, Shewry_Hey_2015}, and consumption is expected to increase in the near future \cite{Erenstein_Jaleta_Mottaleb_Sonder_Donovan_Braun_2022}. The US Great Plains, the focus of this study, produce approximately 6\% of  global production, and 11\% of global traded wheat \cite{usda_ers_glance}. Moreover, the region has an untapped potential that ranges from 36\% to 58\% given biophysical conditions, which is largely attributed to nitrogen \cite{Jaenisch_Munaro_Bastos_Moraes_Lin_Lollato_2021, Giordano_Sadras_Correndo_Lollato_2024} and disease management \cite{Jaenisch_de_Oliveira_2019, Jaenisch_Munaro_Bastos_Moraes_Lin_Lollato_2021, Cruppe_DeWolf_2021, Giordano_Sadras_2025}. Improving wheat productivity relies on integrating and enhancing access to data generated in agronomic research programs.

Data from agronomic field experiments are abundant, with limited integration with data from other sources. This suggests an enormous gap for enhancing interoperability between datasets from varying sources. Here, ontologies can play a significant role in bridging the semantic gap, enabling interoperability between various datasets and platforms, and allowing for scalable, Artificial Intelligence (AI)-driven applications, as seen in other areas of agricultural research \cite{munir2024agrodss, rani2021ontagri, lacasta2021framework}. While ontologies and KGs have been used to describe crop traits and phenotypes for wheat breeding \cite{nedellec2020wto,zhao2024wheatchain}, data from agronomic research trials, and environmental descriptors (soils, weather) remain fragmented. 

Farm management practices, along with key environmental conditions like rainfall, temperature, and soil water holding capacity, are the major factors that can alter the yield in wheat production, thereby either increasing or decreasing the \textit{yield gap} ($Y_G$), which is the difference between the actual crop yield ($Y_a$) and the potential yield that could be achieved under optimal conditions. This metric highlights the potential for increasing crop productivity and the importance of management in achieving this potential. Adopting advanced agronomic practices has shown promise in narrowing the yield gap. Specifically, nitrogen (N) management and foliar fungicide applications have demonstrated significant potential in this regard \cite{jaenisch2022modulation}. Furthermore, a combination of agronomic strategies, including adjustments to sowing date, planting density, and fertilization regimes, has proven effective in reducing the yield disparity \cite{yao2021effects}. Fertilization strongly impacts the wheat yield, although the effect depends heavily on these same environmental factors \cite{bryan2014influence}. In light of these considerations, we identified nutrient management and disease management as two critical modules to be represented in our proposed datahub. Omitting them would overlook two of the most yield-determinant domains within the wheat production ecosystem. While our KG will ultimately support a wide range of agricultural applications, the integration of data it achieves is especially crucial for building a robust framework that addresses one of the most important challenges in modern agriculture, namely closing the yield gap.

Here, we present the Sustainable Wheat Production Datahub, a modular KG framework for wheat production that models nutrient and disease management as federated ontologies joined by a bridging layer. The bridging layer is the small set of shared elements that allows the two domains to work together while each remains separate. It consists of the cross-domain relationships that link a concept in one domain to a concept in the other. The Datahub was built following three objectives:
\begin{enumerate}
    \item Design a federated ontology architecture with two independent modules (one for nitrogen and one for disease management), that interact by sharing concepts and agronomically meaningful interactions (e.g., how disease management is affected by nitrogen nutrition). 
    \item  Assemble a populated KG that brings data from field trials, expert knowledge, and regional climate and drought records into a single, semantically consistent repository. 
    \item  Showcase the KG application to answer agronomically relevant questions pertained to nitrogen and disease management.   
\end{enumerate}

Together, these contributions support downstream AI applications in agriculture and provide a reusable foundation for a broader, semantically rich global food systems datahub.


\section{Related Work}
\label{sec:related_work}
In recent years, KGs have gained popularity due to their numerous benefits, including scalability and organizing heterogeneous data into a single structure, making it easier to adapt in diverse domains \cite{DBLP:journals/cacm/Hitzler21,hogan2021knowledge}. It has emerged as a powerful paradigm for integrating and managing diverse data sources at scale. It is particularly relevant for agricultural data integration, where diverse datasets---ranging from soil health and climate conditions to genetic information and wheat market trends---need to be combined into a unified framework. In agriculture, several efforts have been made to generate a vocabulary that defines standard terms and relationships in the agricultural domain. Agroportal \cite{agroportal}, for example, serves as a hub that contains a number of vocabularies and ontologies related to agriculture. Table~\ref{tab:kg-comparison} contrasts these systems, together with Sustainable Wheat Production Datahub, including a set of architectural dimensions.

Similarly, AgroLD \cite{larmande2025agrold} integrates plant science data from many sources into a single KG of more than 1 billion RDF triples, and Planteome \cite{cooper2024planteome} provides a set of reference and crop-specific ontologies together with a knowledgebase of annotated plant data.

There also exist other KG-centric efforts that are relevant to the goal of creating a global food systems datahub \cite{gfsMcGinty24}. \citet{brewster2024data} propose a data-sharing architecture, the Ploutos, that is based on three principles: a) reuse of existing semantic standards; b) integration with legacy systems; and c) a distributed architecture where stakeholders control access to their own data. The Ploutos semantic model is built on an integration of existing ontologies, which use graph query patterns to traverse the network and collect the requisite data to be shared. Their data-sharing approach is highly extensible, with considerable potential for capturing sustainability-related data, related to global food systems, which includes food, environment, transportation, sensor related data.

More recent efforts extend this perspective toward interoperable and reusable agricultural data. \citet{top2022fair} examine the operationalization of the Findable, Accessible, Interoperable, Reusable (FAIR) principles for agri-food data; \citet{pakrashi2023cowmesh} unify heterogeneous farm data through an ontology-based data mesh; and \citet{bergier2026agritrust} combine an ontology with a federated governance model for trusted data exchange. In the wheat domain, the Wheat Information System \cite{sen2020wheatis} offers a single portal for discovering genomic and phenotypic datasets distributed across international nodes, functioning as a federated data-discovery service rather than a single ontology-driven KG.

\begin{table*}[htbp]
\centering
\caption{Comparison of Sustainable Wheat Production Datahub with representative agricultural (OAK, C3PO, AgroLD, CropDP-KG, MaCOnto), geospatial (KnowWhereGraph), and climate (LinkClimate) knowledge graph (KG) systems along architectural dimensions. Entries reflect what each system reports; ``--'' denotes a capability not reported.}
\label{tab:kg-comparison}
\setlength{\tabcolsep}{4pt}
\resizebox{\textwidth}{!}{%
\begin{tabular}{@{}lllccll@{}}
\toprule
System & Domain focus & Representation & Modular/federated & Multi-domain & Reasoning & Empirical population \\
\midrule
OAK \cite{ngo2020oak}                   & Digital agriculture (broad)  & RDF/OWL              & Yes & Yes & --                      & No (schema) \\
KnowWhereGraph \cite{SHIMIZU2025100842} & Geospatial/environmental     & RDF/OWL, GeoSPARQL   & Yes & Yes & GeoSPARQL               & Yes (30 layers) \\
LinkClimate \cite{wu2022linkclimate}    & Climate (+geo, web)          & RDF                  & No  & Yes & --                      & Yes (NOAA/OSM/Wikidata) \\
C3PO \cite{darnala2023c3po}             & Crop (vegetable) production  & RDF/OWL              & Yes & Yes & DL consistency          & Yes ($\sim$8.4M triples) \\
AgroLD \cite{larmande2025agrold}        & Plant science (omics/traits) & RDF                  & Yes & Yes & --                      & Yes ($\sim$1.08B triples) \\
CropDP-KG \cite{yan2025cropdpkg}        & Crop disease \& pest         & Neo4j (prop.\ graph) & No  & No  & --                      & Yes ($\sim$13.8k entities) \\
MaCOnto \cite{aminu2022maconto}         & Maize nutrient/soil          & OWL + SWRL           & No  & No  & SWRL rules              & Not stated \\
\midrule
\textbf{Sustainable Wheat Production (this work)}               & Wheat nutrient + disease     & RDF/OWL              & Yes & Yes & OWL~2~RL (materialized) & Yes (trials/weather/drought) \\
\bottomrule
\end{tabular}%
}
\end{table*}

Recently, significant efforts have been directed toward the development of domain-specific vocabularies, which are designed to address questions within a particular domain by encompassing only the relevant concepts and relationships. For instance, the Crop Disease Ontology \cite{CDO} provides a hierarchical framework for describing crop health, pathogens, and their impacts on agricultural productivity. Similarly, the Environment Ontology (ENVO) \cite{buttigieg2016environment} defines environmental entities, systems, and processes across diverse domains, supporting applications in ecology, climate studies, and environmental sciences.

Domain-specific ontologies have also been developed for individual crops and for the two areas this work addresses. For crop production, C3PO \cite{darnala2023c3po} organizes a crop planning and production ontology into modules that back an integrating KG, while ICASA \cite{white2013icasa} standardizes the description of field experiments and production data. For nutrient management, MaCOnto \cite{aminu2022maconto} models the soils, fertilizers, and irrigation of the maize crop, and \citet{kessler2021nfert} present an ontology-based decision support system for the nitrogen fertilization of winter wheat. For disease management, CropDP-KG \cite{yan2025cropdpkg} builds a KG of crop diseases and pests, and the EPPO ontology \cite{ayllonbenitez2023eppo} represents plant and pest codes for phytosanitary use.

Likewise, BIMERR Weather Ontology \cite{bimerr}, and Ontology for Meteorological Sensors \citep{meteorologicalscience} define weather-related concepts, observations, and measurements captured by meteorological sensors. These ontologies enable the integration of weather information into intelligent systems, supporting applications in weather prediction, environmental monitoring, and climate research.

Broader KGs integrate data across several themes at once. OAK \cite{ngo2020oak} builds an ontology-based knowledge map for digital agriculture that links soil, weather, crop, and disease information; KnowWhereGraph \cite{SHIMIZU2025100842} provides a geospatial KG spanning natural hazards, climate variables, soil, and land cover; and LinkClimate \cite{wu2022linkclimate} harmonizes climate observations with geographic and encyclopedic data through a climate ontology. Each of these systems remains organized around its own thematic scope, whether agricultural, geospatial, or climatic.

The ontologies mentioned above, along with many other existing ontologies, face limitations in their integration with knowledge from other domains, as they are primarily designed to address specific problems within well-defined domain boundaries. Recent surveys of agricultural KGs likewise point to cross-domain knowledge fusion and federated platforms as open challenges \cite{yang2025grainkg}. Our objective is to address this limitation by developing an integrated framework that describes a wheat production value chain. This proposed datahub will aim to incorporate knowledge spanning the entire lifecycle of wheat, from farm to table.

\section{Methodology}
\label{sec:methodology}
This section describes the methodology used to construct the Sustainable Wheat Production Datahub. The work follows the Knowledge Acquisition and Representation Methodology (KNARM) \cite{mcginty2018knowledge}, an ontology engineering approach that addresses the knowledge acquisition bottleneck by keeping domain experts in the loop throughout development and by orienting the design toward the intended end applications and the available datasets. Within this methodology, we model the two domains that most strongly govern the wheat yield gap, namely nutrient management and disease management. The remainder of the section opens with an overview of the overall architecture (Section~\ref{sec:architecture}) and then proceeds through the phases of KNARM.

\subsection{Architecture Overview}
\label{sec:architecture}
We organize the two domains as a federation of ontologies rather than a single monolithic KG. Each domain is maintained as a distinct ontology, with its own modules and validation cycle. This federated design was adopted for three reasons. First, modularity constrains each domain to a scope that can be independently developed, audited, and reasoned over, which was essential given that the two domains were constructed and refined on separate timelines. Second, a federation permits each module to be reused in isolation, allowing, for example, a study concerned solely with nitrogen dynamics to import the relevant nutrient modules without incorporating the disease schema. Third, separating the domains avoids the maintenance burden of a single large graph in which unrelated concepts become entangled. Rather than fusing the ontologies, the relationships and the observation construct that connect the two domains are placed in a dedicated bridging tier, described next.

\begin{figure*}[t]
  \centering
  \includegraphics[width=0.8\textwidth]{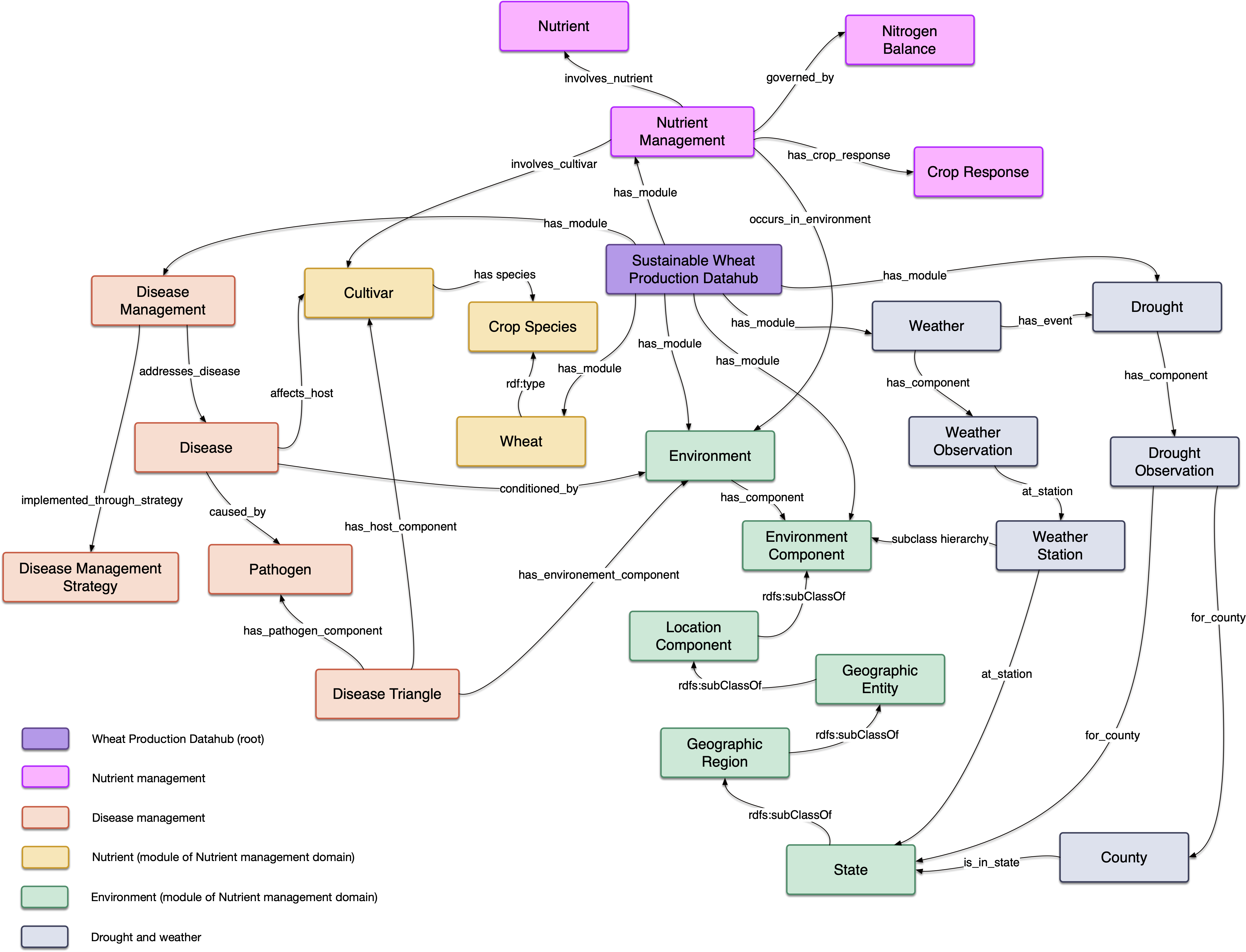}
  \caption{High-level concept map of the Sustainable Wheat Production Datahub.}
  \label{fig:overall}
  \caption*{\footnotesize The figure presents only a small portion of the complete graph, illustrating the interconnections among the modules and the concepts they share, rather than the complete content of any single module.}
  \label{fig:concept-map}
\end{figure*}

\begin{figure*}[htbp]
  \centering
  \includegraphics[width=0.62\textwidth]{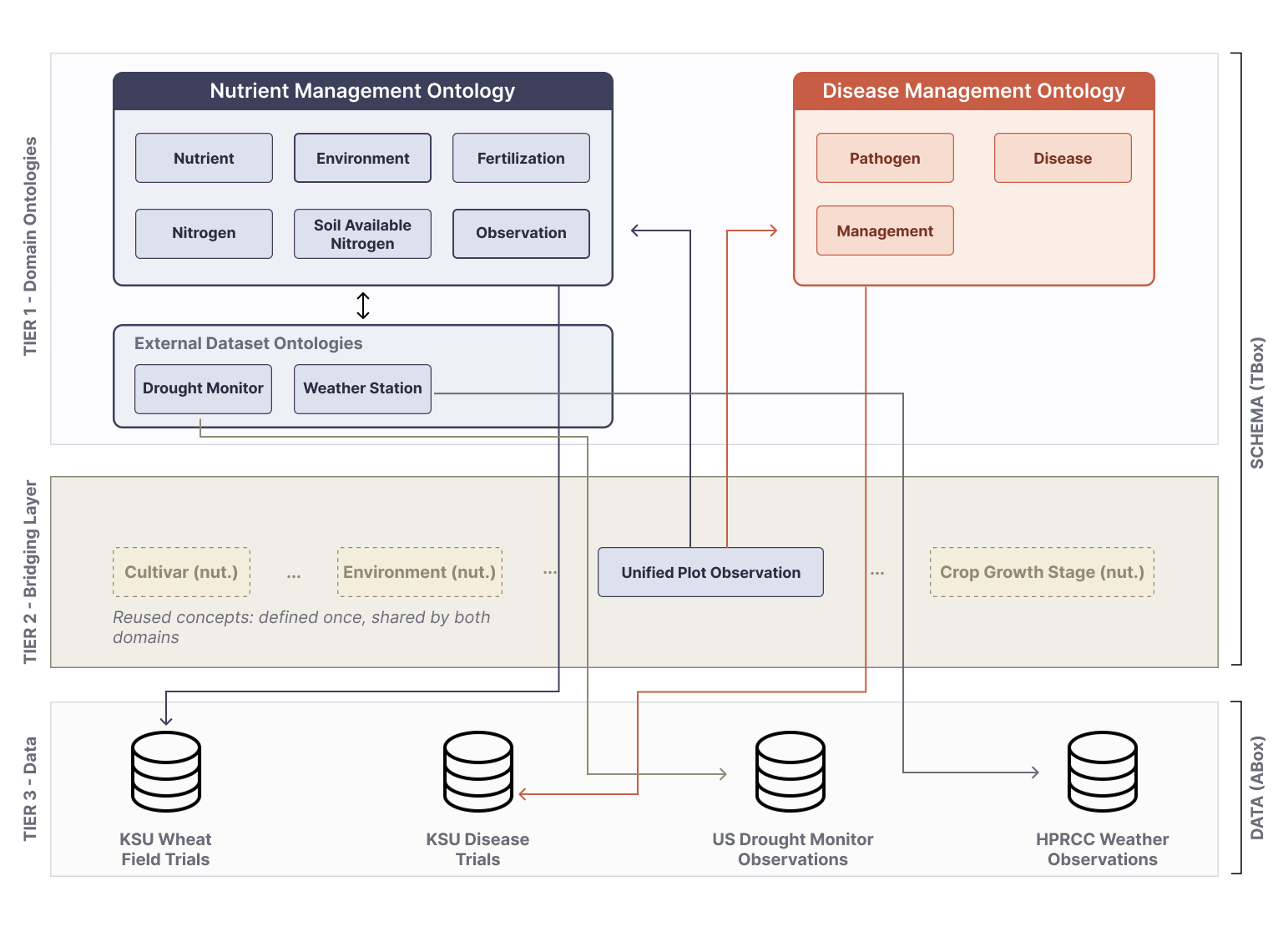}
  \caption{Three-tier architecture of the Sustainable Wheat Production datahub, separating schema (TBox) from data (ABox).}
  \label{fig:architecture}
    \caption*{\footnotesize Three-tier architecture of Sustainable Wheat Production Datahub, which separates the ontology schema (TBox) from the populated data (ABox). Tier 1 holds the two domain ontologies, nutrient management and disease management, with the external dataset ontologies for drought and weather. Tier 2 is a bridging layer that joins two domains, along with a Unified Plot Observation that ties measurements from both domains to a single field plot. Tier 3 holds the data, drawn from Kansas State University (KSU) wheat field trials, disease trials, United States (US) Drought Monitor observations, and High Plains Regional Climate Center (HPRCC) weather observations.}
\end{figure*}

The architecture is organized into three tiers, illustrated in Figure~\ref{fig:architecture}. Tier 1 comprises the two domain ontologies, the nutrient management and disease management modules, which capture the domain-specific knowledge of each area. Tier 2 is a bridging layer that connects the two domains. It comprises two kinds of elements: cross-domain relationships that link a concept in one domain to a concept in the other, e.g., the influence of nitrogen rate on disease susceptibility; and a unified plot observation that binds measurements from both domains to a single field plot. Concepts common to both domains, such as Cultivar, Environment, and Crop Growth Stage, are not collected into a separate layer but are reused directly across modules, defined once and referenced wherever they are needed rather than duplicated. This reuse combines manual modeling with automated assembly: the knowledge engineer decides which module owns each concept and references it elsewhere by its identifier rather than redefining it, and a script then merges the modules into one schema graph, removing the redundant references so that each concept keeps a single definition (Section~\ref{sec:data_population_kg_building}). Tier 3 is the data layer, which holds the populated instances, including field trials, climate records, and drought observations. A concept-level view of this integration is shown in Figure~\ref{fig:overall}, which illustrates how the modules connect and where domains meet at shared concepts.

The schema and the data are maintained separately. Each ontology's classes and properties are maintained under a schema namespace, while the instances that populate them are maintained under a parallel data namespace, distinguished by a -data suffix. This is the standard TBox and ABox separation, and it permits the schema to remain stable while the data grow, allows the two to be versioned independently, and allows a single schema to serve several datasets. The deployment of these namespaces as named graphs, together with the full naming convention, is described in Section~\ref{sec:data_population_kg_building}.

Having described the architecture, we now describe how each ontology was built. The construction followed the nine steps of the Knowledge Acquisition and Representation Methodology (KNARM) \cite{mcginty2018knowledge}, which we group into three phases. The first phase, ontology modeling, comprises the first six steps: \textit{Sub-Language Analysis}, \textit{Unstructured Interviews}, and \textit{Sub-Language Recycling}, which acquire and reuse existing knowledge; \textit{Metadata Creation and Knowledge Modeling}, which formalizes the schema; and \textit{Structured Interview} and \textit{Knowledge Acquisition Validation}, which refine it with domain experts. The second phase, data population and KG building, comprises the seventh and eighth steps, \textit{Database Formation} and \textit{Semi-Automated Ontology Building}, which store the data separately from the schema and populate the graph. The third phase, validation and evaluation, is the ninth and final step, \textit{Ontology Validation and Evaluation}, which assesses the graph for logical soundness and functional adequacy. The remainder of this section presents these three phases.

\subsection{Ontology Modeling}
In ontology modeling, the schema of each domain is designed: the classes, properties, restrictions, and hierarchy that define what the data can express. This phase begins by understanding and reusing existing knowledge (Section~\ref{sec:knowledge_Acquisition_and_reuse}) and by fixing the conventions that govern how every module is modeled (Section~\ref{sec:modeling_conventions_and_quality_criteria}). On that basis, the two domain ontologies, nutrient management and disease management, are built (Sections~\ref{sec:nutrient_management_module} and \ref{sec:disease_management_module}), and the resulting schema is refined through structured review with domain experts before data population (Section~\ref{sec:schema_refinement_and_expert_validation}).

\subsubsection{Knowledge Acquisition and Reuse}
\label{sec:knowledge_Acquisition_and_reuse}
The first three steps of KNARM, namely Sub-Language Analysis, Unstructured Interviews, and Sub-Language Recycling, share a single aim: to understand and reuse the knowledge that already exists in the domain rather than to model it from scratch. Reuse reduces redundancy and improves interoperability, since terms drawn from established sources carry meanings that other systems already recognize. The outcome of these steps is a grounded understanding of the semantic elements and relationships present in the wheat production data ecosystem, together with a set of vocabularies on which the schema can be built.

\textbf{\textit{Sub-Language Analysis}} examines the available data sources, identifies the patterns within them, and characterizes the relationships among them. We reviewed the kinds of data relevant to wheat production, spanning soil characteristics, weather, and regional measures, and from these, we identified the recurring entities, attributes, and relationships that the ontology would need to represent. The analysis surfaced dependency chains that later shaped the schema, e.g., how fertilization acts through the environment and soil characteristics to affect the crop.

\textbf{\textit{Unstructured Interviews}} complement the data analysis with expert knowledge. A group of domain experts\footnote{The term Domain Experts, used throughout this manuscript, refers to researchers and wheat extension specialists in the Department of Agronomy, Kansas State University.} with knowledge of wheat production was interviewed informally and iteratively throughout the process. These conversations surfaced knowledge that the data alone did not capture, such as the practical reasoning behind management decisions, and the resulting insights shaped early schema assumptions that were then revised until they expressed the intended meaning.

\textbf{\textit{Sub-Language Recycling}} reuses terms and vocabularies from existing ontologies and resources. We referenced several established ontologies: the KnowWhereGraph \cite{SHIMIZU2025100842}, the Environment Ontology (ENVO) \cite{buttigieg2016environment}, the Phenotype and Trait Ontology (PATO) \cite{gkoutos2018anatomy}, the Unit Ontology (UO) \cite{gkoutos2012units}, and the Crop Disease Ontology (CD) \cite{CDO}. The KnowWhereGraph, to our knowledge the largest public geo-knowledge graph, covers a substantial body of agriculture-relevant data, including soil health, land use, and land cover, and serves as our infrastructure for location-based information. ENVO provides a computer-readable representation of environmental entities and processes, ranging from whole ecosystems to their constituent parts. PATO offers a standardized vocabulary for phenotypic qualities such as properties and attributes, used alongside other ontologies to describe phenotypes. UO supplies units of measurement, several of which we reused directly for our measurement entities. CD provides crop, disease, and pesticide concepts, drawn largely from agriculture in India. These ontologies served as references that informed our modeling decisions and vocabulary choices rather than as components imported directly into the schema. Building on this grounding, we developed a schema tailored to the wheat production setting and to the structure of the available data.

\subsubsection{Modeling Conventions and Quality Criteria}
\label{sec:modeling_conventions_and_quality_criteria}
Before modeling either domain, we fixed a small set of conventions that every module was required to follow, so that the modules would be consistent with one another, readable to both people and tools, and amenable to automated checking. Every class and property carries a human-readable label and a natural-language comment stating its intended meaning, which keeps the schema self-describing, allows documentation to be generated directly from it, and reduces the risk that a term is reused with an unintended meaning. Naming follows a single convention throughout: local identifiers use Title\_Case with underscores between words (for example, Soil\_Component), while the corresponding rdfs:label gives the same term in readable form without underscores (Soil Component), so that the two stay predictably aligned, with rare exceptions retained only for established domain notation such as soil pH.

Beyond documentation and naming, further conventions structure the schema for reasoning and modularity. Every property declares an explicit domain and range, which fixes the entities it may connect, permits a reasoner to flag assertions that violate that structure, and supplies the information needed to support inference rather than mere storage. We also distinguish classes from individuals consistently: open-ended categories are modeled as classes, whereas fixed, enumerable value sets, such as a defined rating scale or a known list of categories, are modeled as named individuals, keeping the hierarchy focused on genuine types. Finally, each module declares only the terms it owns and refers to sibling terms through lightweight stubs rather than redefining them, so it can be opened, audited, and reasoned over on its own, while a single master module imports all of a domain's modules to assemble the full ontology when needed. Together, these conventions address pitfalls common in ontologies, such as missing annotations, untyped properties, and tangled module dependencies, and set the quality baseline for the two domain ontologies described next.

\subsubsection{Nutrient Management Ontology}
\label{sec:nutrient_management_module}
The Nutrient Management ontology is organized into six modules, unified by a master module that imports them to form the complete domain ontology. The Nutrient module defines the plant nutrients and the crop responses used to evaluate their management; the Environment module describes the soil and weather context in which nutrients act; the Fertilization module represents fertilizer products and the practices through which nutrients are applied; the Nitrogen and Soil Available Nitrogen modules elaborate the nitrogen cycle and the plant-available pool it feeds; and the Observation module binds field measurements to the plots on which they were recorded. An umbrella concept, Nutrient Management, sits in the parent namespace and connects these modules, since every nutrient management program targets at least one nutrient and is evaluated through at least one crop response.

\begin{figure*}[htbp]
  \centering
  \includegraphics[width=0.8\textwidth]{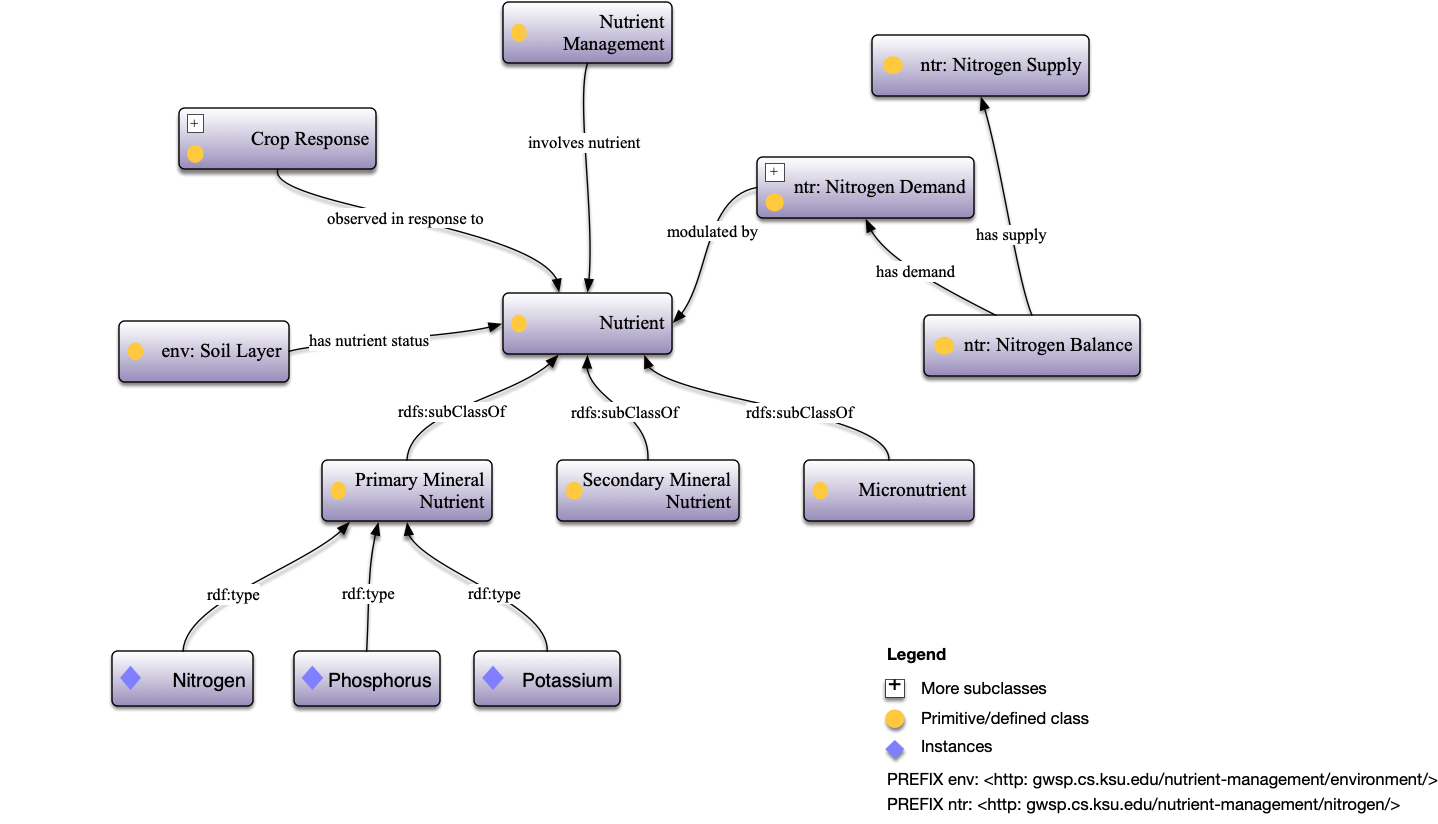}
  \caption{Excerpt of Nutrient Management Ontology}
  \label{fig:nutrient-management}
    \caption*{\footnotesize Excerpt from the Nutrient Management ontology, showing selected concepts rather than the full module. The root Nutrient class divides into primary mineral, secondary mineral, and micronutrient categories through \texttt{rdfs:subClassOf}, while the individual elements, such as nitrogen, phosphorus, and potassium, are modeled as instances via \texttt{rdf:type} rather than as subclasses. A few surrounding concepts, including crop response, soil layer, and nitrogen balance, indicate how a nutrient links to the wider module.}
\end{figure*}

At the core of the domain is the nutrient hierarchy, shown in Figure~\ref{fig:nutrient-management}. A root Nutrient class is divided into primary minerals, secondary minerals, and micronutrients, while the structural elements obtained from air and water (carbon, hydrogen, and oxygen) are deliberately excluded, since the ontology is focused on management through soil and fertilizer inputs. The individual elements, such as nitrogen, phosphorus, and potassium, are modeled as named individuals of these classes rather than as subclasses, in keeping with the class and individual distinction described in Section~\ref{sec:modeling_conventions_and_quality_criteria}: the categories of the nutrient are open types, whereas the elements form a fixed, enumerable set. This keeps the nutrient hierarchy compact, defining 16 classes and 59 individuals, and a parallel crop response hierarchy captures the variables measured to evaluate treatments, grouped into yield, growth, and quality responses, the last including wheat-specific milling and baking measures.

Nitrogen receives the most detailed treatment, reflecting its central influence on wheat yield. Beyond its place as an individual in the nutrient hierarchy, it is elaborated in a dedicated module built around a Nitrogen Balance concept that serves as a central accounting framework linking nitrogen supply, demand, transformations, and losses, while a separate Soil Available Nitrogen module models the plant-available pool that this balance feeds. Together, these modules represent both the field-level nitrogen cycle and the plant-internal dynamics that determine how absorbed nitrogen is partitioned, giving the ontology the depth needed to reason about nitrogen management rather than merely record it. The Observation module then ties this schema to field data: a Plot Observation represents a single plot in a single trial and binds together the treatments applied to it, the cultivar grown, the site and year, and the measured responses such as yield and grain protein, and it is the same construct that the bridging tier designates as the unified plot observation, where nutrient and disease measurements meet on a common plot.
 
\subsubsection{Disease Management Ontology}
\label{sec:disease_management_module}
The Disease Management ontology is organized into three modules, unified by a master module that imports them and defines the root concept Disease Management. The Pathogen module describes the biological agents that cause disease, classified as fungal, bacterial, or viral, together with the reservoirs in which they persist between seasons, whether in soil, crop residue, seed, a living host, or an insect vector, and the mechanisms by which they spread. The Disease module defines the diseases themselves and the conditions under which they develop. The Management module represents the strategies used to control them. The root Disease Management concept connects the three modules, integrating knowledge of the pathogen, the disease, and the available control strategies within the constraints of the environment and the host.

\begin{figure*}[htbp]
  \centering
  \includegraphics[width=0.7\textwidth]{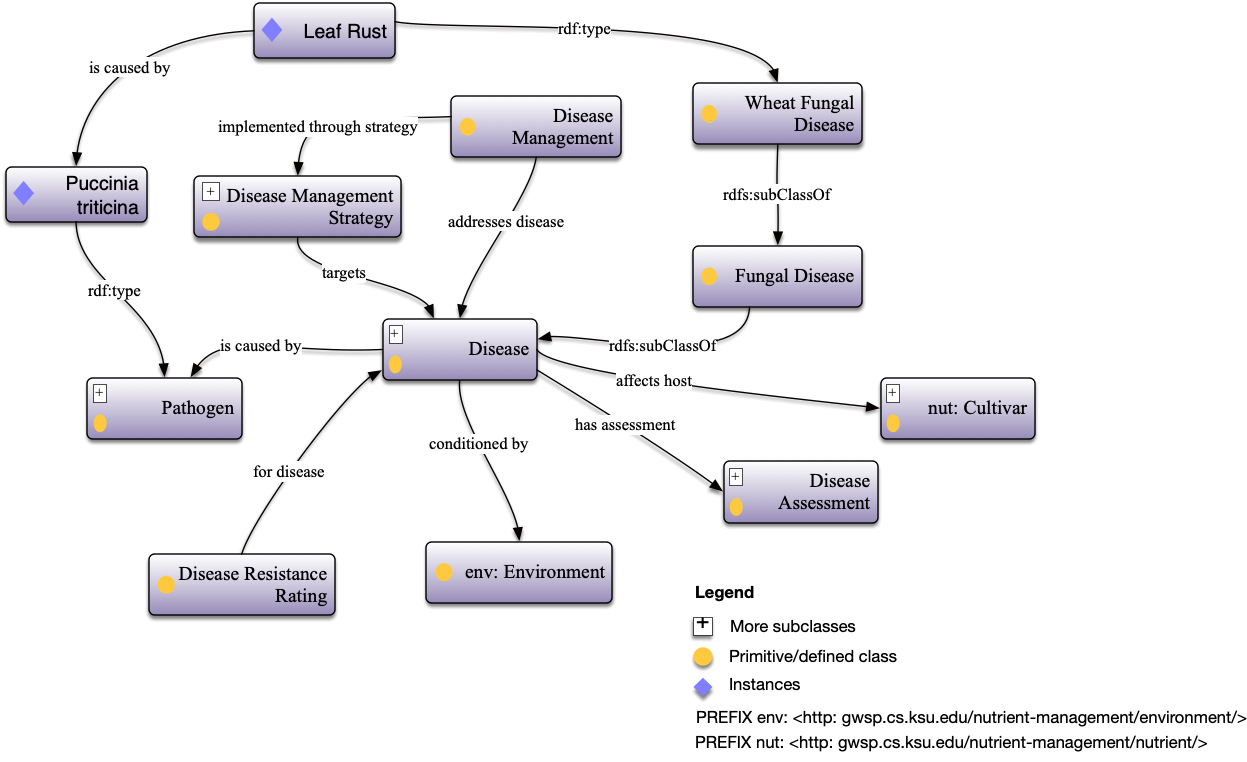}
  \caption{Excerpt of Disease Management Ontology}
  \label{fig:disease-management}
    \caption*{\footnotesize Excerpt from the disease management ontology, showing selected concepts rather than the full module. A Disease is kept distinct from the Pathogen that causes it, illustrated by the instance Leaf Rust, typed as a Wheat Fungal Disease, and its causal agent \texttt{\textit{Puccinia triticina}}, an instance of Pathogen. The Disease class carries the relationships of the disease triangle, being caused by a pathogen, affecting a host cultivar, and conditioned by an environment, and is further linked to a disease assessment and a resistance rating, while a disease management strategy targets it. The host and environment concepts are reused from the Nutrient Management ontology, shown by the \texttt{nut:} and \texttt{env:} prefixes. Yellow markers denote classes and blue markers denote instances, as given in the legend.}
\end{figure*}

A central modeling decision separates a disease from the pathogen that causes it, shown in Figure~\ref{fig:disease-management}. Leaf rust is a disease, whereas \textit{Puccinia triticina} is the pathogen responsible for it, and keeping the two distinct allows the same pathogen to be associated with different diseases under different conditions while keeping pathogen biology and disease symptoms in their proper places. Each disease is modeled through the disease triangle, the standard plant pathology model in which disease arises from the interaction of a susceptible host, a virulent pathogen, and a favorable environment \cite{francl2001disease,scholthof2007disease,leach2024reenvisioning}. The Disease class, therefore, carries restrictions stating that a disease is caused by a pathogen, affects a host cultivar, is conditioned by an environment, and is evaluated through a disease assessment. The Host and Environment vertices reuse the Cultivar and Environment concepts from the Nutrient Management ontology, so the disease triangle is one of the points at which the two domains already meet. Diseases divide into fungal, bacterial, and viral categories, and the ten major fungal diseases of wheat, such as leaf rust, stem rust, and stripe rust, are modeled as named individuals linked to their causal organisms, with bacterial and viral diseases represented in the same way, again following the class and individual distinction of Section~\ref{sec:modeling_conventions_and_quality_criteria}. Disease assessment captures incidence, the proportion of plants or plots that are infected, and severity, the extent to which each infected plant is affected, and disease-specific resistance ratings record how a given cultivar responds to a given disease.

The Management module models integrated disease management as a set of strategies, each disrupting one or more vertices of the disease triangle. Three pillars are represented: chemical control, which applies fungicides classified by their mode-of-action group (FRAC)\footnote{\url{https://www.frac.info/}}, active ingredient, and application timing; cultural control, the agronomic practices that reduce disease pressure; and host resistance, the genetic control achieved through cultivar selection, distinguishing vertical from horizontal resistance. Framing each strategy as the disruption of a disease triangle vertex makes the rationale for a control measure explicit and supports reasoning about which combination of measures suits a given situation. The cultural practices reuse the management practice concept from the Nutrient Management ontology for the specific operations that serve both domains, such as managing crop residue through tillage, which affects the soil nitrogen supply while also reducing the inoculum that survives between seasons. This reuse applies only where a practice genuinely serves both purposes. With both domain ontologies modeled, the final modeling steps refine and validate the schema before any data are populated.

\subsubsection{Schema Refinement and Expert Validation}
\label{sec:schema_refinement_and_expert_validation}
The modeling described in the preceding subsections constitutes the fourth step of KNARM, \textbf{\textit{Metadata Creation and Knowledge Modeling}}, in which the concepts, their attributes, and the relationships among them are formalized into the module schemas. Two further steps then refine and check that schema before any data are populated. The fifth step, \textbf{\textit{Structured Interview}}, complements the earlier unstructured interviews with close-ended questions posed to the same domain experts, used to confirm specific modeling choices and to resolve points left open during the informal discussions. At this stage, we also formulated a set of competency questions, the concrete information needs that the ontology is expected to answer. These questions serve as a completeness benchmark, since a schema that can answer them is judged to cover the intended scope. The competency questions, their translation into SPARQL, and the results of executing them against the populated graph are presented in Section~\ref{sec:results_and_evaluation}.

The sixth step, \textbf{\textit{Knowledge Acquisition Validation}}, is the first feedback loop in the process. The knowledge engineer presents the outputs of the earlier steps, including the reused vocabularies, the metadata, and the assembled schema, back to the domain experts, whose review identifies concepts that were missed, misinterpreted, or modeled at the wrong level of detail. The schema is revised in response, and the cycle repeats until the experts agree that it represents the domain faithfully. Through this iterative refinement, the schema reaches a stable form configured to support data population, which is the subject of the next section.

\subsection{Data Population/KG Building}
\label{sec:data_population_kg_building}
With a validated schema in place, the second phase of KNARM populates it with data. This phase covers two steps: creating a database that holds the data separately from the schema, and building the populated KG in a semi-automated manner.

The seventh step, \textbf{\textit{Database Formation}}, establishes a storage structure for the data that is kept distinct from the schema. The database may be relational, graph-based, or document-oriented. We use GraphDB 10.8.2\footnote{\url{https://graphdb.ontotext.com/}}, a graph-native triple store, for its structural compatibility with an RDF schema, its scalability to large graphs, and its native support for SPARQL. The repository is configured with an OWL 2 RL rule set, so that facts entailed by the schema are computed and stored at the time of data loading, and are then available to queries directly, without running a separate reasoning step at query time.

The data are deployed in the store as a set of named graphs that keep each dataset's schema apart from its instances. A fully populated dataset occupies three graphs. The schema graph holds the TBox, namely the classes, properties, restrictions, and comments, and for a domain ontology, this single graph gathers the terms of all that domain's modules. The reference graph holds the stable anchor individuals to which observations point, such as locations, trials, cultivars, states, and weather stations, which change only when new entities are introduced. The observations graph holds the bulk assertional data, with one individual per source record, and is the part that grows and is refreshed as new source data are published. Each schema graph is named for the source or model it formalizes, while the two data graphs share a {domain}-data prefix, so that the role of any graph is readable from its name. Table~\ref{tab:namespaces} lists the resulting convention. The deployed datahub currently comprises twelve named graphs, three each for the Nutrient Management ontology, the Disease Management ontology, the drought dataset, and the weather dataset. Holding schema, reference individuals, and bulk observations in separate graphs allows the observations to be refreshed without disturbing the validated schema, permits the schema and each dataset be versioned independently, and facilitates each layer be queried on its own.

\begin{table*}[t]
\centering
\caption{Named-graph convention for the Sustainable Wheat Production Datahub, a modular,
ontology-driven knowledge graph. Each dataset separates its schema (TBox) from a reference
graph of stable anchor individuals and an observation graph of bulk per-record
individuals. All paths share the base \texttt{\url{https://gwsp.cs.ksu.edu/}}.}
\label{tab:namespaces}
\footnotesize
\begin{tabular}{@{}llll@{}}
\toprule
Dataset & Schema (TBox) & Reference (Stable ABox) & Observations (Bulk ABox) \\
\midrule
Nutrient management & \texttt{nutrient-management/schema} & \texttt{nutrient-data/reference} & \texttt{nutrient-data/observations} \\
Disease management & \texttt{disease-management/schema} & \texttt{disease-data/reference} & \texttt{disease-data/observations} \\
Drought             & \texttt{drought-monitor/schema}     & \texttt{drought-data/reference}  & \texttt{drought-data/observations} \\
Weather             & \texttt{weather-station/schema}     & \texttt{weather-data/reference}  & \texttt{weather-data/observations} \\
\bottomrule
\end{tabular}
\end{table*}

Several datasets were curated, cleaned, and ingested following this convention. The primary source is the set of wheat crop performance field trials\footnote{\href{https://www.agronomy.k-state.edu/outreach-and-services/crop-performance-tests/wheat/}{K-State Wheat Crop Performance Tests}} conducted at Kansas State University, which records, for each plot, the cultivar grown, the management treatments applied, and the measured responses such as yield, grain protein, and test weight. Drought severity data from the U.S. Drought Monitor\footnote{\url{https://droughtmonitor.unl.edu/}} provides weekly observations of drought area, categorized from D0 (abnormally dry) to D4 (exceptional drought), across the central Great Plains states. Daily climate data from the High Plains Regional Climate Center (HPRCC)\footnote{\url{https://hprcc.unl.edu/}} supplies station-level temperature, precipitation, and snow measurements over a multi-decade period. These quantitative sources are complemented by qualitative expert insights, including those gathered at a stakeholder sustainability workshop, which inform the representation of sustainable practices in the graph \cite{giordano2026fragmented}.

The bulk of the population work converts the tabular field-trial data into RDF instances typed against the schema. Each row of the trial spreadsheet becomes a Plot Observation individual in the observations graph, linked to the cultivar, the treatments, site, and year that describe it, and carrying its measured values as datatype properties. These instance files contain no schema definitions; they reference the schema classes and properties solely by their identifiers, which is the TBox and ABox separation realized in practice. Several modeling choices from the observation module are applied during this conversion. The three treatment slots recorded in the data are each represented through the same Treatment Factor pattern, so that fertility, variety, and timing treatments are handled uniformly. The most recent previous crop is stored as the actual crop grown, such as soybean or canola, and linked to a broad crop category, so that both the specific crop and its agronomic class can be queried. Missing measurements, which are common in field data, are handled by omitting the corresponding triple rather than inserting a null, so that the graph asserts only what was actually observed. The stable entities that many observations share, namely locations, years, trials, cultivars, and site-year environments, are created once in the reference graph and referred to by the observations, which avoids duplication and ensures that anchors remain consistent.

The eighth step, \textbf{\textit{Semi-Automated Ontology Building}}, assembles and populates the graph using established tooling. We use the RDFLib\footnote{\url{https://github.com/RDFLib/rdflib}} library in Python to construct the schema modules and to generate the instance data programmatically. The approach is semi-automated because the construction is scripted while domain experts remain engaged throughout to confirm that the result is accurate and relevant. The per-module schema files are merged into a single consolidated graph, with a deduplication pass that removes redundant declarations introduced by cross-module references, and this consolidated graph is loaded as the schema graph. Files are loaded in formats matched to their size, with the compact Turtle serialization used for the small schema and reference graphs, and a streaming-friendly serialization used for the large observation graphs.

\subsection{Validation and Evaluation}
\label{sec:validation_and_evaluation}
Once a stable version of the ontology is produced, it is validated along two complementary lines: logical soundness, which assesses whether the ontology is internally consistent and classifies as intended, and functional adequacy, which evaluates whether the ontology can answer the questions it was built to answer. Knowledge engineers and domain experts carry out these checks together, the former responsible for the formal properties of the schema and the latter for the faithfulness of its content. \textbf{\textit{Ontology Validation and Evaluation}}, the final step of KNARM, brings together the logical and functional checks described above.

Logical soundness is assessed with an OWL 2 reasoners. We use two fully conformant OWL 2 DL reasoners, HermiT and Pellet \citep{glimm2014hermit, sirin2007pellet}, to confirm that the schema is consistent, that no class is unsatisfiable, and that the inferred class hierarchy matches the intended one, which surfaces any unintended subsumptions introduced during modeling. This check is distinct from the rule-based materialization applied in the triple store, which computes query-time entailments under the OWL 2 RL profile. The schema is intentionally restricted to an ontology fragment that supports efficient reuse by downstream AI applications. By limiting constructs such as disjointness and negation, the design reduces the likelihood of logical inconsistencies by construction. Reasoner-based validation, therefore, primarily verifies correct classification, detects modeling errors as the schema evolves, and is repeated whenever new axioms are introduced.

Functional adequacy is assessed through competency questions based on concrete information needs formulated with domain experts. Each question is translated into a SPARQL query and run against the populated graph, and a schema that returns correct and complete answers is judged to cover its intended scope. Validation is iterative rather than a single terminal step, since expert feedback and new data may introduce further axioms that require re-checking. The competency questions (CQ) and their results are presented in Section~\ref{sec:results_and_evaluation}.

\section{Results and Evaluation}
\label{sec:results_and_evaluation}

\subsection{Overview}
This section evaluates the Sustainable Wheat Production datahub along the two dimensions introduced in Section~\ref{sec:validation_and_evaluation}: logical soundness, namely whether the ontology is internally consistent and classifies as intended, and functional adequacy, namely whether it satisfies the information needs for which it was designed. In addition, we demonstrate that the KG supports inference beyond retrieval, one of the capabilities that distinguishes a graph-based representation from simpler data structures.

The evaluation comprises four parts. First, we establish logical soundness and report the structural statistics of the constructed KGs (Section~\ref{sec:logical_validation_and_knowledge_graph_statistics}). Second, we assess functional adequacy for the Nutrient Management domain through competency questions, formulated in plain language, translated into SPARQL, and executed against the populated graph (Section~\ref{sec:competency_questions_nutrient_management}). Third, we assess functional adequacy for the Disease Management domain in the same manner (Section~\ref{sec:competency_questions_disease_management}). Fourth, we demonstrate reasoning through queries whose answers derive from inferred rather than asserted facts (Section~\ref{sec:reasoning_over_the_knowledge_graph}). Both domains are fully modeled, logically validated, and populated with field-trial data. The Nutrient Management domain is exercised over the crop performance trials, and the Disease Management domain over the disease trials, which record fungicide applications and disease assessments, so both domains are evaluated over their populated data as well as their modeled concepts.

\subsection{Logical Validation and Knowledge Graph Statistics}
\label{sec:logical_validation_and_knowledge_graph_statistics}

Logical validation was performed with fully conformant OWL 2 DL reasoners. Each domain ontology was loaded with all of its modules and classified. Both the Nutrient Management and the Disease Management ontologies were found to be logically consistent, and no class was determined to be unsatisfiable. This check is repeated whenever new axioms are introduced, so that consistency is preserved as the schema evolves.

The composition of the two ontologies is summarized in Table~\ref{tab:schema-stats}, which reports, for each domain, the number of classes, object properties, datatype properties, and named individuals after the modules are merged. The counts reflect the class and individual discipline described in Section~\ref{sec:modeling_conventions_and_quality_criteria}: elements and enumerable value sets are represented as named individuals rather than as classes, which keeps the class counts compact relative to the number of named entities the ontology defines.

The scale of the populated graphs is reported in Table~\ref{tab:kg-stats}, which provides the number of RDF triples in each named graph of the deployment described in Section~\ref{sec:data_population_kg_building}. The schema graphs are small and human-authored; the nutrient field-trial observations graph holds the per-plot data generated from the trial records; and the external climate and drought KGs account for the large majority of the triples in the datahub.

\begin{table}[t]
\centering
\caption{Composition of the two domain ontologies of Sustainable Wheat Production Datahub, nutrient management and disease management, counted from the merged schema of each domain.}
\label{tab:schema-stats}
\footnotesize
\begin{tabular}{@{}lcc@{}}
\toprule
 & Nutrient Management & Disease Management \\
\midrule
Classes              & 149 & 45 \\
Object properties    & 66 & 28 \\
Datatype properties  & 35 & 10 \\
Named individuals    & 197 & 58 \\
Logical axioms & 708 & 189 \\
\bottomrule
\end{tabular}
\end{table}

\begin{table}[t]
\centering
\caption{RDF triple counts for each named graph in the deployment of Sustainable Wheat Production Datahub}
\label{tab:kg-stats}
\footnotesize
\begin{tabular}{@{}lr@{}}
\toprule
Named graph & \# Triples \\
\midrule
nutrient-management/schema      & 3{,}042 \\
nutrient-data/reference         & 458 \\
nutrient-data/observations      & 53{,}950 \\
disease-management/schema       & 643 \\
disease-data/reference          & 44 \\
disease-data/observations       & 12{,}663 \\
drought-monitor/schema          & 2{,}227 \\
drought-data/reference          & 3{,}768 \\
drought-data/observations       & 10{,}731{,}500 \\
weather-station/schema          & 71 \\
weather-data/reference          & 858 \\
weather-data/observations       & 6{,}355{,}300 \\
\bottomrule
\end{tabular}
\end{table}

\subsection{Competency Questions: Nutrient Management}
\label{sec:competency_questions_nutrient_management}
We formulated 20 competency questions for the nutrient management domain. They span two levels: schema-level questions that interrogate the modeled concepts, such as the classification of nutrients and the structure of the nitrogen balance, and data-level questions that draw on the populated field-trial data, such as yield and grain protein. Each was translated into a SPARQL query and executed against the populated graph, and the schema returned an answer to every question. The complete set, with each query and its result, is provided in the project repository\footnote{\url{https://github.com/koncordantlab/Sustainable-Wheat-Production-Datahub/blob/main/competency-questions/COMPETENCY_QUESTIONS.md}}. Here we present two worked examples that illustrate the range of queries the ontology supports: a semantic aggregation that follows the crop-category links, and a multi-attribute join across treatment, environment, and measurement.

The first example asks for the average yield of the following wheat crop for each previous crop category. The query joins each plot observation to the crop grown in the preceding season and to that crop's agronomic category, then averages the recorded yield within each category (Listing~\ref{lst:CQ_N16}, Table~\ref{tab:yield_by_prevcrop}). Because previous crops are stored as the actual crop and linked to a broad category such as legumes or cereals (Section~\ref{sec:data_population_kg_building}), the question is answered at the category level even though the data record only the specific crop. This traversal, from an observation to a crop to its category, is the kind of semantic link the graph resolves directly.

\begin{lstlisting}[style=sparqlstyle, caption=SPARQL query for average wheat yield by previous-crop category (CQ N16)., label={lst:CQ_N16}]
PREFIX rdfs: <http://www.w3.org/2000/01/rdf-schema#>
PREFIX obs:  <https://gwsp.cs.ksu.edu/nutrient-management/observation/>
PREFIX fert: <https://gwsp.cs.ksu.edu/nutrient-management/fertilization/>
SELECT ?cat (SAMPLE(?catLabel) AS ?category)
       (AVG(?y) AS ?avgYield) (COUNT(?o) AS ?n) 
WHERE {
  ?o a obs:Plot_Observation ;
     obs:observed_previous_crop ?pc ;
     obs:has_yield_lbs_per_acre_13pct ?y .
  ?pc fert:has_crop_category ?cat .
  OPTIONAL { ?cat rdfs:label ?catLabel }
}
GROUP BY ?cat
ORDER BY DESC(?avgYield)
\end{lstlisting}

\begin{table}[htbp]
    \centering
    \caption{Average wheat yield by previous-crop category, obtained by running the SPARQL query for Competency Question N16 against the populated nutrient management data in Sustainable Wheat Production Datahub. Yields are reported in pounds per acre (lbs/acre), the unit recorded in the source field-trial data; for reference, 1{,}000 lbs/acre is approximately 1.12 t/ha. Obs.\ ($n$) is the number of observations averaged in each category.}
    \label{tab:yield_by_prevcrop}
    \begin{tabular}{l @{\hspace{10pt}} c @{\hspace{10pt}} r}
        \toprule
        \textbf{Category} & \textbf{Avg. Yield (lbs/acre)} & \textbf{Obs. ($n$)} \\
        \midrule
        Fallow   & 6014 & 64 \\
        Oilseeds & 5655 & 287 \\
        Cereals  & 4042 & 377 \\
        Legumes  & 3887 & 1647 \\
        \bottomrule
    \end{tabular}
\end{table}

The second example asks, for each variety and location, the maximum observed grain protein and test weight, ranked by protein. The query links each observation to its cultivar and, through the site-year environment, to its location, then takes the maximum of each measurement within every variety-location pair and ranks the pairs by protein (Listing~\ref{lst:CQ_N11}). The highest-ranked combinations are shown in Table~\ref{tab:top_varieties_protein_testweight}. A single query thus combines a treatment factor, an environmental context, and two grain-quality measurements, joined through the plot observation that binds them.

\begin{lstlisting}[style=sparqlstyle, caption=SPARQL query for top variety-location combinations by grain protein (CQ N11)., label={lst:CQ_N11}]
PREFIX rdfs: <http://www.w3.org/2000/01/rdf-schema#>
PREFIX obs:  <https://gwsp.cs.ksu.edu/nutrient-management/observation/>
SELECT ?variety ?location
       (MAX(?protein) AS ?maxProtein) (MAX(?tw) AS ?maxTestWeight) 
WHERE {
  ?o a obs:Plot_Observation ;
     obs:observed_cultivar     ?cv ;
     obs:observed_at_site_year ?sye ;
     obs:has_protein_pct_dry_basis  ?protein ;
     obs:has_test_weight_lbs_per_bu ?tw .
  ?cv  rdfs:label ?variety .
  ?sye obs:has_location ?loc .
  ?loc rdfs:label ?location .
}
GROUP BY ?variety ?location
ORDER BY DESC(?maxProtein)
LIMIT 50
\end{lstlisting}

\begin{table*}[htbp]
    \centering
    \caption{Top variety–location combinations by grain protein, obtained by running the SPARQL query for Competency Question N11 against the populated nutrient management data in Sustainable Wheat Production Datahub.}
    \label{tab:top_varieties_protein_testweight}
    \begin{tabular}{lccr}
        \toprule
        \textbf{Variety} & \textbf{Location} & \textbf{Protein (\% dry basis)} & \textbf{Test Weight (lbs/bu)} \\
        \midrule
        GreenHammer   & Great Bend       & 18.5 & 57.45 \\
        WB4458        & Ashland Bottoms  & 17.8 & 59.90 \\
        WB4303        & Ashland Bottoms  & 17.6 & 58.21 \\
        WB Grainfield & Ashland Bottoms  & 17.5 & 60.10 \\
        WB4515        & Ashland Bottoms  & 17.4 & 57.70 \\
        \dots        & \dots            & \dots & \dots \\
        \bottomrule
    \end{tabular}
\end{table*}

Together, these two examples show the ontology answering questions that range from a semantic aggregation over thousands of plot records to a multi-hop join that unites treatment, environment, and measurement. The same pattern extends to the remaining questions, which are documented in full in the project repository.

\subsection{Competency Questions: Disease Management}
\label{sec:competency_questions_disease_management}
We formulated 20 competency questions for the disease management domain. They span two levels: schema-level questions that interrogate the modeled concepts, such as the classification of pathogens and the structure of the disease triangle, and data-level questions that draw on the populated disease trial data, such as fungicide applications and recorded disease severity. Each was translated into a SPARQL query and executed against the populated graph, and the graph provided an answer to every question. The complete set, with each query and its result, is provided in the project repository. Here we present two worked examples: a schema-level query that exercises the separation of a disease from its causal pathogen, and a data-level query that joins fungicide applications to disease assessments across the trial plots.

The first example asks, for each wheat disease, which pathogen causes it. The query follows the \texttt{caused\_by} relation from every disease to its causal organism (Listing~\ref{lst:CQ_D8}, Table~\ref{tab:disease_pathogen}). Because diseases and pathogens are modeled as distinct entities (Section~\ref{sec:disease_management_module}), the mapping is explicit and queryable, and a single pathogen genus can appear for more than one disease, as with the two Fusarium diseases caused by different Fusarium species. This is the type of distinction that a model conflating disease and pathogen could not express.

\begin{lstlisting}[style=sparqlstyle, caption=SPARQL query for the causal pathogen of each wheat disease (CQ D3)., label={lst:CQ_D8}]
PREFIX rdfs: <http://www.w3.org/2000/01/rdf-schema#>
PREFIX dis:  <https://gwsp.cs.ksu.edu/disease-management/disease/>
SELECT ?disease ?diseaseLabel ?pathogen ?pathogenLabel WHERE {
  ?disease dis:caused_by ?pathogen .
  OPTIONAL { ?disease  rdfs:label ?diseaseLabel }
  OPTIONAL { ?pathogen rdfs:label ?pathogenLabel }
}
ORDER BY ?diseaseLabel
\end{lstlisting}

\begin{table}[htbp]
    \centering
    \caption{Causal pathogen of each wheat disease, obtained by running the SPARQL query for Competency Question D3 against the disease management ontology in Sustainable Wheat Production Datahub.}
    \label{tab:disease_pathogen}
    \begin{tabular}{l @{\hspace{20pt}} r}
        \toprule
        \textbf{Disease} & \textbf{Causal Pathogen} \\
        \midrule
        Bacterial Leaf Streak            & \textit{Xanthomonas translucens} \\
        Barley Yellow Dwarf          & \textit{Barley Yellow Dwarf Virus} \\
        Common Bunt  & \textit{Tilletia spp.} \\
        Fusarium Foot Rot & \textit{Fusarium spp.} \\
        Fusarium Head Blight   & \textit{Fusarium graminearum} \\
        \dots                 & \dots \\
        \bottomrule
    \end{tabular}
\end{table}

The second example draws on the populated disease trial data. It compares treated and untreated plots on disease pressure, mycotoxin, and yield, where a plot counts as treated if it received at least one fungicide application, and peak severity is the maximum severity recorded across a plot's assessments. Only Fusarium Head Blight produces deoxynivalenol, so for the other diseases this comparison rests on severity alone; we keep the mycotoxin column here for the sake of the example. The query joins each plot to its fungicide applications and to its disease assessments at once (Listing~\ref{lst:CQ_D13}, Table~\ref{tab:cq-d13}), a join that a flat trial table cannot express. This is a direct demonstration of the value of the populated graph: the treatment status, the disease outcome, and the yield are held on separate records and brought together only through the plot that binds them.

\begin{lstlisting}[style=sparqlstyle, caption={SPARQL query comparing treated and untreated plots on disease, mycotoxin, and yield (CQ D13).},label={lst:CQ_D13}]
PREFIX obs: <https://gwsp.cs.ksu.edu/nutrient-management/observation/>
PREFIX dis: <https://gwsp.cs.ksu.edu/disease-management/disease/>
PREFIX mgt: <https://gwsp.cs.ksu.edu/disease-management/management/>
SELECT ?treatment (COUNT(?plot) AS ?plots)
       (AVG(?peakSeverity) AS ?meanPeakSeverity)
       (AVG(?don)   AS ?meanDON)
       (AVG(?yield) AS ?meanYield)
       (AVG(?tw)    AS ?meanTestWeight)
WHERE {
  {
    SELECT ?plot (MAX(?sev) AS ?peakSeverity) WHERE {
      ?a dis:assessed_plot ?plot ;
         dis:has_severity_percent ?sev .
    }
    GROUP BY ?plot
  }
  OPTIONAL { ?plot dis:has_DON_ppm ?don }
  OPTIONAL { ?plot obs:has_yield_bu_per_acre ?yield }
  OPTIONAL { ?plot obs:has_test_weight_lbs_per_bu ?tw }
  BIND( IF( EXISTS { ?app mgt:applied_to_plot ?plot }, "treated", "untreated" )
        AS ?treatment )
}
GROUP BY ?treatment
\end{lstlisting}

\begin{table*}[htbp]
    \centering
    \caption{Treated versus untreated plots on peak disease severity, deoxynivalenol (DON), yield, and test weight, obtained by running the SPARQL query for Competency Question D13 against the populated disease trial data in Sustainable Wheat Production Datahub.}
    \label{tab:cq-d13}
    \begin{tabular}{l @{\hspace{10pt}} c @{\hspace{10pt}} c @{\hspace{10pt}} c @{\hspace{10pt}} c @{\hspace{10pt}} r}
        \toprule
        \textbf{Treatment} & \textbf{Plots} & \textbf{Mean peak sev. (\%)} & \textbf{Mean DON (ppm)} & \textbf{Mean yield (bu/ac)} & \textbf{Mean test wt. (lbs/bu)} \\
        \midrule
        Treated   & 204 & 30.38 & 15.64 & 74.74 & 46.13 \\
        Untreated & 20 & 53.04 & 27.09 & 56.24 & 38.45 \\
        \bottomrule
    \end{tabular}
\end{table*}

Together, these two examples show the graph answering questions of two kinds: a schema-level mapping that traverses the disease-to-pathogen relation, and a data-level join over the trial records that brings treatment, disease outcome, and yield onto a common plot. The same pattern extends to the remaining questions, which are documented in full in the project repository.

\subsection{Cross-Source Querying}
Wheat yield is shaped not only by management but by the environmental conditions of each season, so a datahub built to study the yield gap must be able to relate management outcomes to the conditions under which they occurred. Because every dataset is expressed against the shared schema and held in the same store, a single query can draw on more than one of them at once. We illustrate this by relating the field-trial yields to the weather records: the query computes the mean wheat yield for each harvest season from the nutrient observations, and the mean growing-season precipitation across Kansas weather stations for the same season, returning them together (Listing~\ref{lst:fed_query}, Table~\ref{tab:cross_source}). The two datasets share no field identifier, yet they are joined through the crop season and restricted to Kansas using the same state entity that the weather stations, the drought records, and the nutrient environment module all refer to. This is the operational payoff of the federated design: independently sourced datasets remain jointly queryable through the shared schema.

\begin{lstlisting}[style=sparqlstyle, caption=SPARQL query for the mean wheat yield and Kansas growing-season precipitation by harvest year., label={lst:fed_query}]
PREFIX rdfs: <http://www.w3.org/2000/01/rdf-schema#>
PREFIX xsd:  <http://www.w3.org/2001/XMLSchema#>
PREFIX obs:  <https://gwsp.cs.ksu.edu/nutrient-management/observation/>
PREFIX env:  <https://gwsp.cs.ksu.edu/nutrient-management/environment/>
PREFIX wx:   <https://gwsp.cs.ksu.edu/weather-station/>

SELECT ?harvestYear
       (AVG(?yield) AS ?avgYield)
       (AVG(?seasonPrecip) AS ?avgKansasSeasonPrecip)
WHERE {
  {
    SELECT ?harvestYear (AVG(?y) AS ?yield) 
    WHERE {
      ?o a obs:Plot_Observation ;
         obs:observed_at_site_year ?sye ;
         obs:has_yield_lbs_per_acre_13pct ?y .
      ?sye obs:has_year ?yr .
      ?yr  rdfs:label ?yrLabel .
      BIND( xsd:integer( SUBSTR( REPLACE(STR(?yrLabel), "[^0-9]", ""), 1, 4 ) ) + 1
            AS ?harvestYear )
    }
    GROUP BY ?harvestYear
  }

  {
    SELECT ?harvestYear ?station (SUM(?precip) AS ?seasonPrecip) 
    WHERE {
      ?w a wx:Weather_Observation ;
         wx:at_station ?station ;
         wx:has_observation_date ?date ;
         wx:has_precipitation_in ?precip .
      ?station env:is_in_state env:Kansas .
      BIND( YEAR(?date) AS ?dy )
      BIND( MONTH(?date) AS ?dm )
      BIND( IF(?dm >= 10, ?dy + 1, ?dy) AS ?harvestYear )
      FILTER( ?dm >= 10 || ?dm <= 6 )
    }
    GROUP BY ?harvestYear ?station
  }
}
GROUP BY ?harvestYear
ORDER BY ?harvestYear
\end{lstlisting}

\begin{table}[htbp]
    \centering
    \caption{Mean wheat yield and Kansas growing-season precipitation by harvest year, retrieved from Sustainable Wheat Production Datahub in a single cross-source query. The yield and precipitation figures come from two independently sourced datasets, the nutrient management field trials and the weather station observations.}
    \label{tab:cross_source}
    \begin{tabular}{l @{\hspace{5pt}} c @{\hspace{5pt}} r}
        \toprule
        \textbf{Harvest year} & \textbf{Mean yield (lbs/acre)} & \textbf{KS. season precip. (in)} \\
        \midrule
        2017 & 4513.5 & 24.1 \\
        2018 & 4108.9 & 14.8 \\
        2019 & 4221.4 & 30.6 \\
        2020 & 4129.2 & 22.0 \\
        \bottomrule
    \end{tabular}
\end{table}

\subsection{Reasoning over the Knowledge Graph}
\label{sec:reasoning_over_the_knowledge_graph}
Beyond retrieving asserted facts, the datahub answers queries whose results are entailed by the schema rather than stored explicitly. Inference is materialized in the triple store under the OWL 2 RL profile (Section~\ref{sec:data_population_kg_building}), so that derived statements are available to queries alongside asserted ones. The clearest case follows from the modeling decision in Section~\ref{sec:nutrient_management_module}, in which each element is represented as an individual of a leaf class such as Primary Mineral Nutrient, and never directly as a Nutrient. The membership of an element in the broad Nutrient class is therefore never asserted; it is derived from the subclass hierarchy. A query posed at the level of Nutrient consequently returns every element, even though no element is stated to be a Nutrient, because the reasoner classifies each one through its leaf class. The same mechanism applies across the schema, allowing questions to be asked at a general level and answered from specific assertions.

This inference is also traceable, which matters for decision support. Each derived fact is produced by a specific rule from specific asserted statements, and GraphDB can report that derivation on request through its proof plugin \citep{graphdb_proof}. A system built on the graph can therefore return an answer and also justify it, rather than presenting an opaque result. Presenting such derivations to end users within a decision-support interface is left to the future work discussed in Section~\ref{sec:conclusion_discussions_and_future_work}, where the explicit path from premises to conclusion serves as the foundation for prescriptive and explainable applications.

\section{Discussion, Future Work, and Conclusion}
\label{sec:conclusion_discussions_and_future_work}
We have presented Sustainable Wheat Production Datahub, a modular KG framework for wheat production that integrates heterogeneous agricultural data within a single ontology-driven structure. Following the KNARM methodology\cite{mcginty2018knowledge} and working with domain experts, we represented the two domains that most strongly govern the wheat yield gap, nutrient management and disease management, as federated ontologies joined by a bridging layer, and populated the graph from empirical field trials together with regional climate and drought records. We validated the ontologies for logical consistency and showed, through competency questions translated into SPARQL, that the graph answers practical agronomic questions across both domains.

Representing these data as a KG, rather than as isolated tables, makes the relationships among concepts explicit and machine-readable. A query can therefore traverse links that a flat dataset cannot follow without manual joins, such as relating a plot to the agronomic category of its previous crop, or a disease to its causal pathogen and to the host and environment that condition it. The formalized schema further supports inference: facts entailed by the class hierarchy are derived rather than stored, so a query can return results that were never explicitly asserted, and each derived fact can be traced to the assertions that justify it (Section~\ref{sec:reasoning_over_the_knowledge_graph}). This combination of explicit relationships, derivation, and traceability is what separates the datahub from simpler structures whose use is limited to retrieval and correlation.

The present work has limitations that define its scope. The integrated data are regional, drawn from the central Great Plains; the external ontologies we reused supplied vocabulary, not data. The populated graph therefore reflects the production systems and growing conditions of that area rather than wheat production more broadly. Expanding the geographic coverage and incorporating a wider range of data sources would strengthen the empirical foundation and allow the framework to support analyses beyond those that the current data permit. Our evaluation establishes logical soundness and functional adequacy through competency questions; a study of the framework within a downstream decision-support application, and a formal assessment with end users, are left to future work. We regard these as boundaries of the current deployment rather than of the approach, which is designed to extend along each of them as further data and domains are incorporated.

Future work will focus first on broadening the data upon which the KG draws. In the near term, this entails full ingestion of publicly available datasets from software like Decision Support System for Agrotechnology Transfer (DSSAT\footnote{\url{https://dssat.net/}}), and Agriculture Model Intercomparison and Improvement Project (AgMIP) \cite{ROSENZWEIG2013166}, and over the longer term, it opens the possibility of incorporating non-public data as well, such as records contributed by individual farms, which would support the graph capture production practices that public datasets do not reach. Furthermore, to preserve the consistency among terms in agriculture, we have envisioned incorporating standard terms from ICASA Data Dictionary\footnote{\url{https://github.com/DSSAT/ICASA-Dictionary}} and others into the KG, beyond the currently implemented datahub. All of these external datasets will act as a module that helps us to preserve the context and scale up the system. Another promising direction is the automated construction and enrichment of KGs using large language models (LLMs), which can extract and structure knowledge from unstructured agronomic texts such as extension manuals, scientific literature, and field reports \cite{zhang2024olive,SHIMIZU2025100862}.

To the best of our knowledge, the Sustainable Wheat Production Datahub framework presented here is the first unified wheat-focused agricultural datahub that semantically integrates heterogeneous datasets within a single ontology-driven framework. Furthermore, the system adheres to FAIR principles, ensuring that its data are Findable, Accessible, Interoperable, and Reusable, thereby maximizing its research and practical utility. Ultimately, we envision this work as a step toward building a global food systems datahub that not only captures the complexity of agricultural systems but also promotes sustainability, equity, and resilience across the food value chain.

\section*{Declaration on Generative AI}
During the preparation of this work, the authors used Claude (Anthropic) in order to assist with: Grammar and spelling check, and improve writing style. After using this tool/service, the authors reviewed and edited the content, and they take full responsibility for the publication's content.

\section*{Supplemental Material}
\sloppy
The GitHub repository, available at
\url{https://github.com/koncordantlab/Sustainable-Wheat-Production-Datahub},
contains the complete schema for both domain ontologies and for the drought and
weather dataset ontologies, the full populated data for each of the four
datasets listed in Table~\ref{tab:namespaces}, and the 40 competency
questions with their SPARQL queries. Browsable documentation for every
ontology, generated with WIDOCO \citep{garijo2017widoco}, is published at
\url{https://gwsp.cs.ksu.edu/ontology/}.

\section*{Acknowledgement}
This work was funded by Kansas State University under the Game-changing Research Initiative Program (GRIP). We also thank Ontotext for providing GraphDB free of charge for this project.



\bibliographystyle{elsarticle-harv}
\bibliography{example}

\end{document}